\documentclass{article}

\usepackage{arxiv}

\usepackage[utf8]{inputenc} 
\usepackage[T1]{fontenc}    
\usepackage{hyperref}       
\usepackage{url}            
\usepackage{booktabs}       
\usepackage{amsfonts}       
\usepackage{nicefrac}       
\usepackage{microtype}      
\usepackage{lipsum}		
\usepackage{graphicx}
\usepackage{natbib}
\usepackage{doi}
\usepackage{subcaption}
\newcommand*\samethanks[1][\value{footnote}]{\footnotemark[#1]}

\title{\emph{Hot-Distance}: Combining One-Hot and Signed Boundary Distance Embeddings for Segmentation}

\date{June 14, 2024}	

\author{ 
    \href{https://orcid.org/0000-0002-9441-2908}{\includegraphics[scale=0.06]{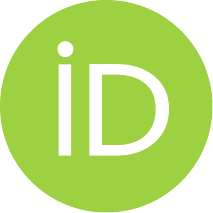}\hspace{1mm}Marwan Zouinkhi}\thanks{contributed equally} \\
	Janelia Research Campus\\
	Howard Hughes Medical Institute\\
	Ashburn, VA 20147 \\
	\texttt{zouinkhim@hhmi.org} \\
	\And
	\href{https://orcid.org/0000-0001-5077-2533}{\includegraphics[scale=0.06]{orcid.pdf}\hspace{1mm}Jeff L. Rhoades}\samethanks \\
	Janelia Research Campus\\
	Howard Hughes Medical Institute\\
	Ashburn, VA 20147 \\
	\texttt{rhoadesj@hhmi.org} \\
    \And
	\href{https://orcid.org/0000-0003-1694-4420}{\includegraphics[scale=0.06]{orcid.pdf}\hspace{1mm}Aubrey V. Weigel}\thanks{corresponding author} \\
	Janelia Research Campus\\
	Howard Hughes Medical Institute\\
	Ashburn, VA 20147 \\
	\texttt{weigela@hhmi.org} \\
}

\renewcommand{\shorttitle}{\textit{Hot-Distance}: Combining One-Hot and Signed Distance Embeddings for Segmentation}

\hypersetup{
pdftitle={HotDistance},
pdfsubject={q-bio.QM, cs.AI, cs.CV, cs.LG, stat.ML, eess.IV},
pdfauthor={Marwan Zouinkhi, Jeff L. Rhoades, Aubrey V. Weigel},
pdfkeywords={computer vision, machine learning, image segmentation, electron microscopy},
}

\begin{document}
\maketitle

\begin{abstract}
	Machine learning models are only as good as the data to which they are fit. As such, it is always preferable to use as much data as possible in training models. What data can be used for fitting a model depends a lot on the formulation of the task. We introduce \emph{Hot-Distance}, a novel segmentation target that incorporates the strength of signed boundary distance prediction with the flexibility of one-hot encoding, to increase the amount of usable training data for segmentation of subcellular structures in focused ion beam scanning electron microscopy (FIB-SEM).

\end{abstract}


\section{Introduction}
Data is the lifeblood of machine learning. It is crucial for practitioners to maximize the accuracy and diversity of data supplied to a model during training in order to achieve a reliable and generalizable final product. The target a model is trained to predict determines what data can be used. For instance, a model designed to produce segmentations may be trained to generate one-hot encoded or signed boundary distance predictions \citep{Heinrich2021-ap}. A network trained to predict the presence of a given structure in terms of a binary mask can be accurately trained on datasets indicating the objects presence, as well as data sparsely representing its absence (see section \ref{sec:onehot}). However, this is not the case for signed boundary distance predictions (see section \ref{sec:signeddistance}). As a result, datasets labeling the presence of structures mutually exclusive to the target objects, but not the targets themselves, cannot be used to train a signed boundary distance prediction model.

We introduce \emph{hot-distance} to incorporate the benefits of signed boundary distance prediction while maintaining the ability to train models on a larger body of datasets. An empirical study of this strategy's effects, compared to existing methods (i.e. one-hot and signed boundary distances) is ongoing and will be published once concluded.

\section{Related Work}
\label{sec:relatedwork}

\subsection{One-hot Encoding}
\label{sec:onehot}

\textit{One-hot} encoding is named for the original representation within a simple circuit, where a single wire in a set is activated (i.e. set to the high voltage for the circuit), and all other wires are set low, indicating that the input corresponds to the output indicated by the "hot" wire. This method of encoding classes has a distinct advantage over encoding classes as an ordinal variable (i.e. 1 for cat, 2 for dog, and 3 for bird) in that it does not imply any relationship between classes that a network would otherwise have to learn to decorrelate. 

For instance, with one-hot encoding, instead of needing to learn to produce a single output close to 2 for an image of a dog and near 1 for a cat, a network can learn to set one output higher than any number of other outputs to indicate a class. This has the additional added advantage that it allows a network to predict class probabilities across multiple object classes simultaneously. This also means that networks can be trained on both \textit{true negative} and \textit{true positive} examples. Consider a picture of multiple animals, which has only been labeled for pixels belonging to dogs. With a one-hot encoding, a network can still be trained to learn what pixels \textit{do not} belong to a cat, based on the presence of the mutually exclusive dog label. Other portions of the image, for which the true label is unknown, simply won’t contribute to the loss calculated for the cat label. This allows partially labeled datasets to be leveraged effectively for training.

\begin{figure}
    \centering
    \subcaptionbox{\centering}{
    \includegraphics[width=0.45\textwidth]{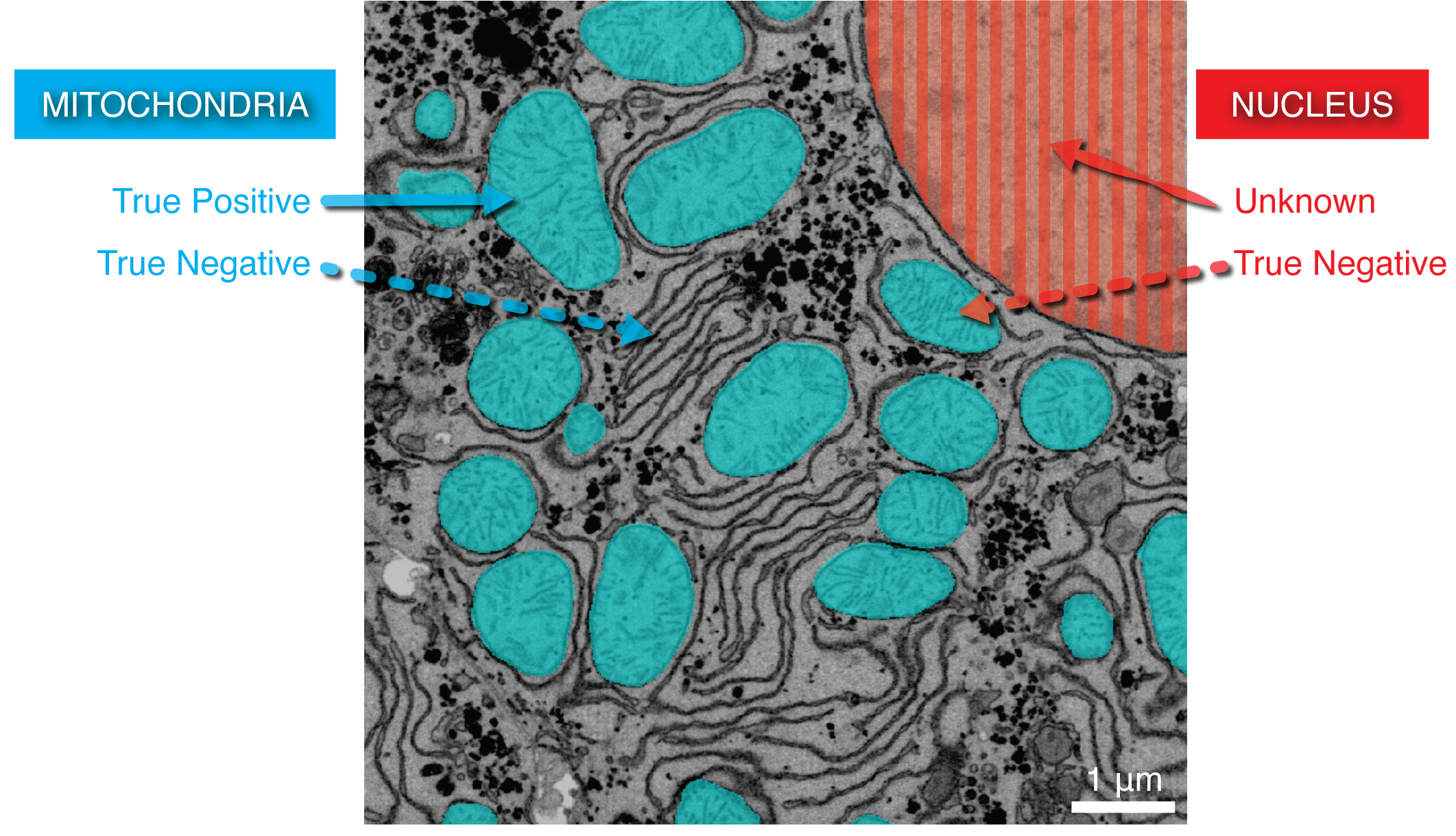}}
    \hspace{0.05\textwidth}
    \subcaptionbox{\centering}{\includegraphics[width=0.45\textwidth]{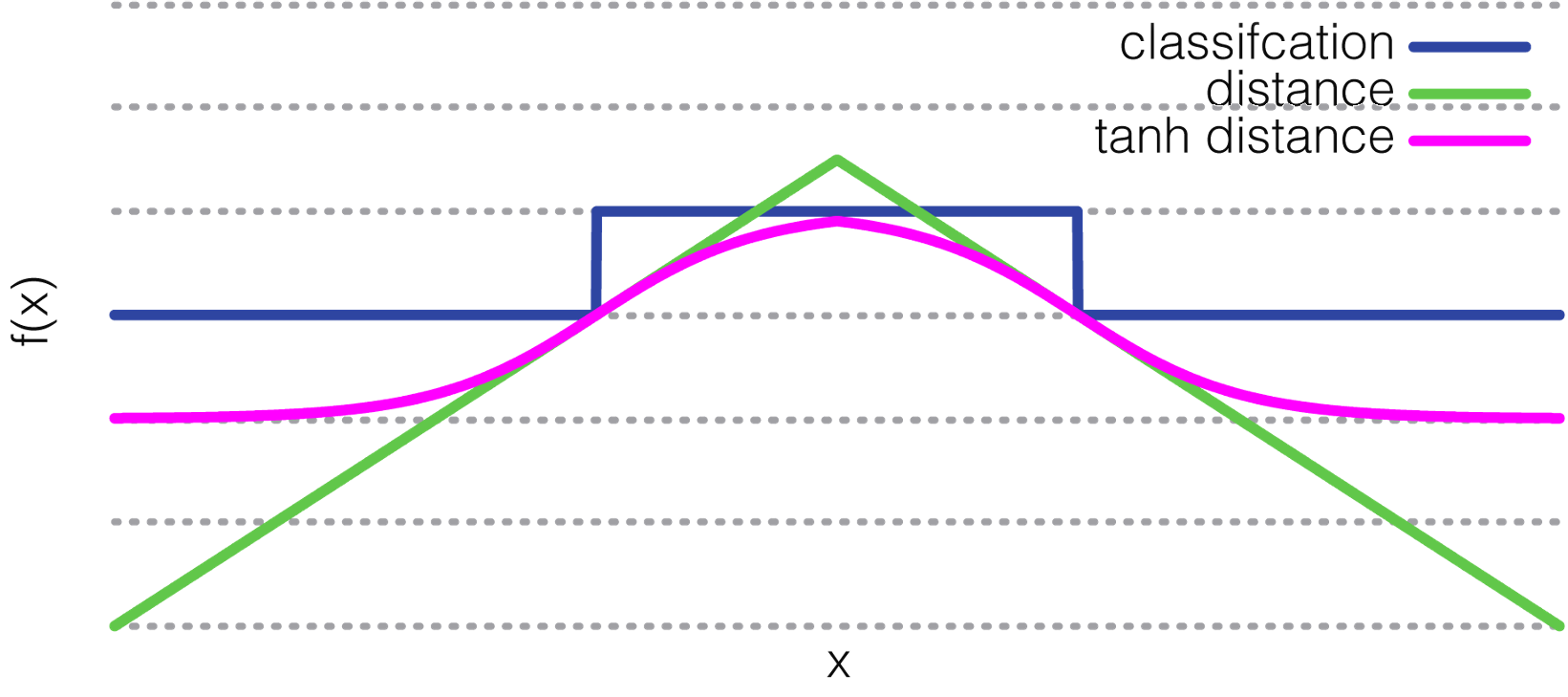}}
    \caption{
    \textbf{Advantages of network prediction targets.}
    (a) One-hot embeddings can represent true negative examples, even in the absence of known positive examples of a class. Here, true negatives are known for the nucleus class (red striped label) based on the true positive examples of the mutually exclusive mitochondria class (blue label).
    (b) Signed tanh boundary distances present smooth gradients for network predictions that can allow for produce instance segmentations via watershed post-processing.
    }
    \label{fig:signeddistance}
\end{figure}

\subsection{Signed boundary distances}
\label{sec:signeddistance}
In contrast to one-hot encoding, which presents a pixel-wise classification task, signed Euclidean boundary distances constitute a pixel-wise regression problem. Pixel-wise regression has been found empirically to outperform the equivalent pixel-wise classification task in object segmentation \citep{Heinrich2018-bb, Kainz2015-og}. Furthermore, it allows for straightforward post-processing into semantic segmentations, via thresholding, or even into instance segmentations via watershed. 

However, the benefit of predicting the output of a smoothly varying monotonic function on class labels may be limited to cases in which informative structures are within the field of view (FOV) of a network. Put another way, a network can predict that a cat is not within a FOV, but it is likely to have a much more difficult time predicting how far away a cat is from its FOV. A formulation of this regression task that has seen great success thus included a hyperbolic tangent function (tanh) as a nonlinearity to bound the distances the network is asked to learn \citep{Heinrich2021-ap, Heinrich2018-bb}. 

Unlike one-hot encodings, networks learning to predict signed boundary distances cannot receive any informative signals in the absence of dense true positive and negative labels. Without dense groundtruth labeling of a class, no information can be provided to a network regarding the distances to positive or negative instances of the label. This greatly reduces the pool of usable training data and increases the barrier to entry for increasing the amount of available data, as everything must be densely annotated.

\section{Hot-Distance}
\label{sec:hotdistance}
In order to incorporate the strengths of both one-hot encodings and signed boundary distance prediction, we introduce \textit{Hot-Distance}. This involves simply predicting both targets with the same network. Splitting of the network signal between the class probability and signed boundary distance predictions occurs only at the last layer of the model, meaning that almost all of the network parameters are shared. Since one-hot predictions can be trained without dense class annotations (for instance, from true negative instances of a class), most network parameters can thus be accurately updated with sparse ground truth data. By leveraging gradients from sparse data, the network backbone can learn to properly identify object features using a much larger set of data.

\begin{table}[h!]
\caption{Comparison of loss functions}
\centering
\begin{tabular}{p{0.25\linewidth} | p{0.2\linewidth} | p{0.2\linewidth}}

\toprule
\textbf{Loss Function} & \textbf{Can use\newline non-target crops} & \textbf{Can learn gradients / do watershed} \\
\midrule
One-hot & \checkmark &  \\
\hline
Signed Boundary Distance &  & \checkmark \\
\hline
Hot-Distance & \checkmark & \checkmark \\
\bottomrule
\end{tabular}
\label{table:loss_functions}
\end{table}




\section{Conclusion}
\label{sec:conclusion}

In this paper, we introduced \textit{Hot-Distance}, a novel approach that combines the strengths of one-hot encoding and signed boundary distance embeddings for the task of segmentation in focused ion beam scanning electron microscopy (FIB-SEM). By leveraging the flexibility of one-hot encoding and the informative gradients of signed boundary distances, \textit{Hot-Distance} enables the use of a broader range of training data, including datasets with sparse annotations.

Our method addresses the limitations of existing approaches by allowing for effective learning from non-target crops and providing the capability to perform gradient-based learning and watershed segmentation. An implementation of this task is available at \href{https://github.com/janelia-cellmap/dacapo/blob/main/dacapo/experiments/tasks/hot_distance_task.py}{github.com/janelia-cellmap/dacapo/blob/main/dacapo/experiments/tasks/hot\_distance\_task.py}

In our next publication, we will present empirical comparison of \textit{Hot-Distance} with traditional one-hot and signed boundary distance loss functions, examining its potential to improve segmentation performance by utilizing a more extensive and diverse training dataset. Future work will focus on the thorough empirical evaluation of \textit{Hot-Distance} across various datasets and segmentation tasks, as well as exploring the integration of additional loss functions to further enhance its performance.

\bibliographystyle{unsrtnat}
\bibliography{paperpile}  

\end{document}